\documentclass{article}

\usepackage[preprint,nonatbib]{neurips_2024}

\usepackage[utf8]{inputenc} 
\usepackage[T1]{fontenc}    
\usepackage{hyperref}       
\usepackage{url}            
\usepackage{booktabs}       
\usepackage{amsfonts}       
\usepackage{nicefrac}       
\usepackage{microtype}      
\usepackage{xcolor}         

\usepackage{enumitem}
\usepackage{graphicx}

\title{Extending Llama-3's Context Ten-Fold Overnight}

\author{Peitian Zhang$^{1,2}$,~ Ninglu Shao$^{1,2}$,~ Zheng Liu$^1$\thanks{Corresponding author.},~ Shitao Xiao$^1$,~ \textbf{Hongjin Qian}$^{1,2}$,\\
        \textbf{Qiwei Ye$^1$,} ~ \textbf{Zhicheng Dou}$^2$\\
        $^1$ Beijing Academy of Artificial Intelligence \\ 
        $^2$ Gaoling School of Artificial Intelligence, Renmin University of China \\ 
        \texttt{namespace.pt@gmail.com} \quad
        \texttt{zhengliu1026@gmail.com} \\
}

\begin{document}

\maketitle

\begin{abstract}
We extend the context length of Llama-3-8B-Instruct from 8K to 80K via QLoRA fine-tuning\footnote{The model is noted as Llama-3-8B-Instruct-80K-QLoRA given its max context length during fine-tuning. However, users could apply the model for even longer contexts via extrapolation.}. The entire training cycle is super efficient, which takes 8 hours on one 8xA800 (80G) GPU machine. The resulted model exhibits superior performances across a broad range of evaluation tasks, such as NIHS, topic retrieval, and long-context language understanding; meanwhile, it also well preserves the original capability over short contexts. The dramatic context extension is mainly attributed to merely 3.5K synthetic training samples generated by GPT-4 , which indicates the LLMs' inherent (yet largely underestimated) potential to extend its original context length. In fact, the context length could be extended far beyond 80K with more computation resources. Therefore, the team will publicly release the entire resources (including data, model, data generation pipeline, training code) so as to facilitate the future research from the community: \url{https://github.com/FlagOpen/FlagEmbedding}. 
\end{abstract}

\section{Introduction}
Recently, considerable attention has been directed towards long-context large language models, where different approaches are adopted to establish long-context capabilities for large language models~\cite{chen2023position_interpolation,peng2023yarn,chen2024longlora,ding2024longrope,fu2024data_engineer,zhang2024soaring,an2024make_your_llm_utilize_context}. However, most of them require significant compute and resources to accomplish.

In this technical report, we propose an efficient solution for entitling the long-context capabilities for LLMs, with which we extend the context length of Llama-3-8B-Instruct\footnote{\url{https://llama.meta.com/llama3/}} from 8K to 80K. Specifically, we use GPT-4~\cite{openai2024gpt4} to synthesize 3.5K long-context training data, covering three long-context tasks:
\begin{enumerate}[leftmargin=20pt,rightmargin=20pt]
    \item \textbf{Single-Detail QA}: the inquiry targets on one specific detail in a long context. To construct data for this task, we slice out a short segment (e.g., a chunk with less than 4096 tokens) from a long context (e.g., a book or a long paper) and prompt GPT-4 to generate multiple question-answer pairs based on this segment.
    \item \textbf{Multi-Detail QA}: the inquiry requires information aggregation and reasoning over multiple details in a long context. We define two types of long context. The \textbf{homogeneous} context contains a coherent text, such as a book or a long paper. We prompt GPT-4 to generate multiple question-answer pairs that require aggregating and analyzing information from different locations in the context. The \textbf{heterogeneous} context consists of multiple independent texts. Notably, we perform clustering over a large corpus then extract texts from the same cluster to form each heterogeneous context. Therefore, the grouped texts share some semantic similarity. We then prompt GPT-4 to ask about the similarities/dissimilarities across these texts.
    \item \textbf{Biography Summarization}: we prompt GPT-4 to write a biography for each main character in a given book.
\end{enumerate}

For all three tasks, the length of context is between 64K to 80K. Note that longer data can also be synthesized following the same methodology. When training, we organize the question-answer pairs for the same context in one multi-turn conversation then fine-tune the LLM to correctly answer the questions given the entire long context as input. 
Following previous work\footnote{https://www.together.ai/blog/llama-2-7b-32k}, we mix 5K instances randomly chosen from RedPajama~\cite{together2023redpajama} to mitigate forgetting. We also mix LongAlpaca~\cite{chen2024longlora} in the training set, which contains 12K instruction tuning instances with 16K length at maximum. Therefore, the entire training dataset contains 20K instances.

We use QLoRA~\cite{dettmers2023qlora} to efficiently fine-tune the model. We apply LoRA on all Q,K,V,O projections and additionally train the embedding layer. We set LoRA rank to 32 and alpha to 16. The learning rate is 5e-5 with linear decay and no warmups. The batch size is 8. Gradient checkpointing is enabled. No parallel strategy is required thanks to the efficient implementation from Unsloth~\cite{unsloth}. We train the model for 1 epoch, which takes 8 hours to complete on a 8xA800 (80G) machine. Importantly, we expand the RoPE base from 500K to 200M in training.

Our contributions are highlighted as follows:
\begin{itemize}[leftmargin=20pt,rightmargin=20pt]
    \item We release Llama-3-8B-Instruct-80K-QLoRA, which extends the context length of Llama-3-8B-Instruct from 8K to 80K. The entire resources including the model, training data, and code are all publicly available, which may advance the field of training long-context LLMs.
    \item Our training recipe is simple and efficient, while the resulted model demonstrates remarkable performance on downstream long-context tasks. Further research can be made to improve our approach.
\end{itemize}

\section{Experiments}
\begin{figure}[t]
    \centering
    \includegraphics[width=\textwidth]{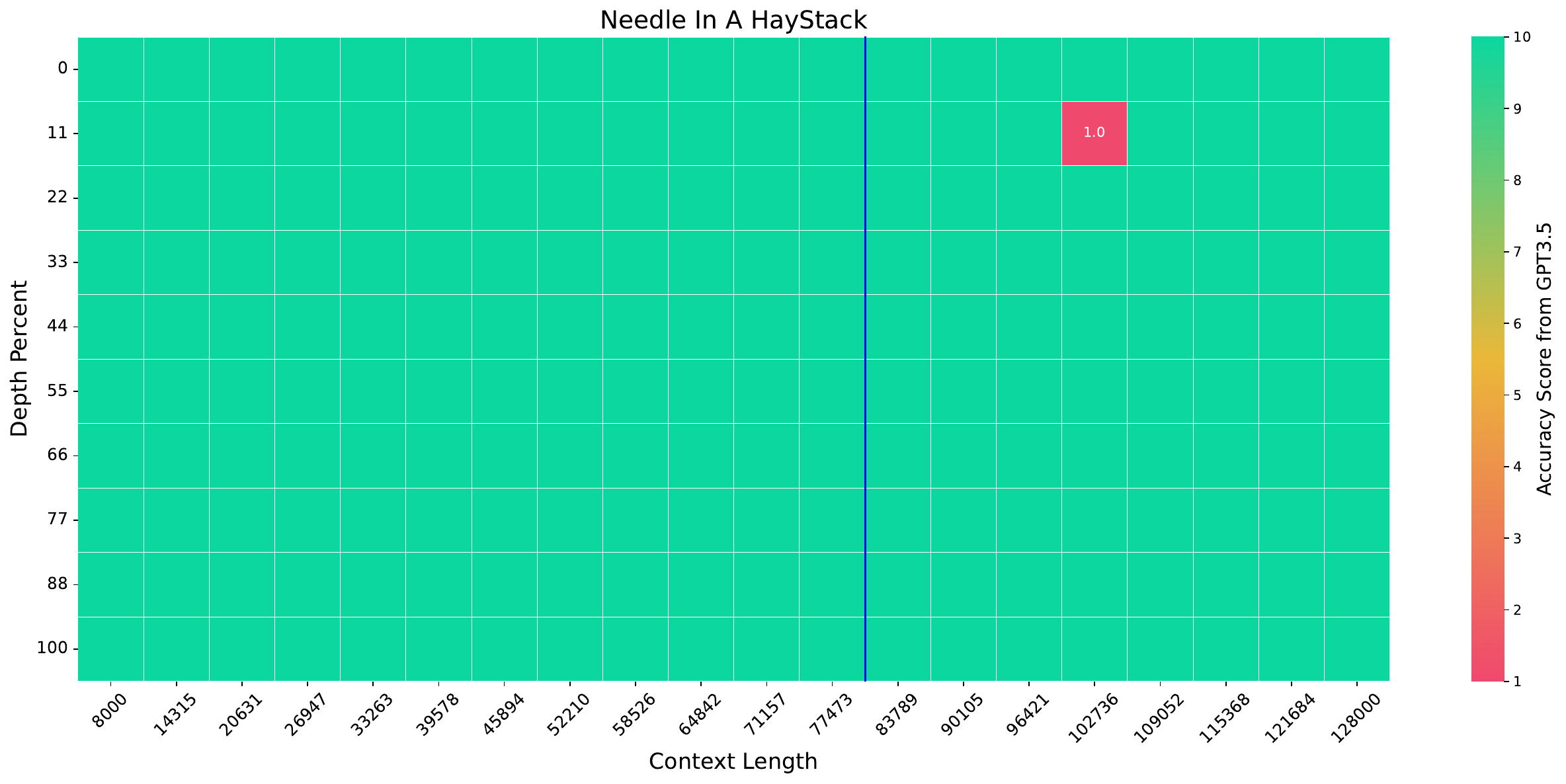}
    \caption{The accuracy score of Llama-3-8B-Instruct-80K-QLoRA on Needle-In-A-HayStack task. The blue vertical line indicates the training length, i.e. 80K.}
    \label{fig:needle}
\end{figure}

\begin{figure}[t]
    \centering
    \includegraphics[width=0.9\textwidth]{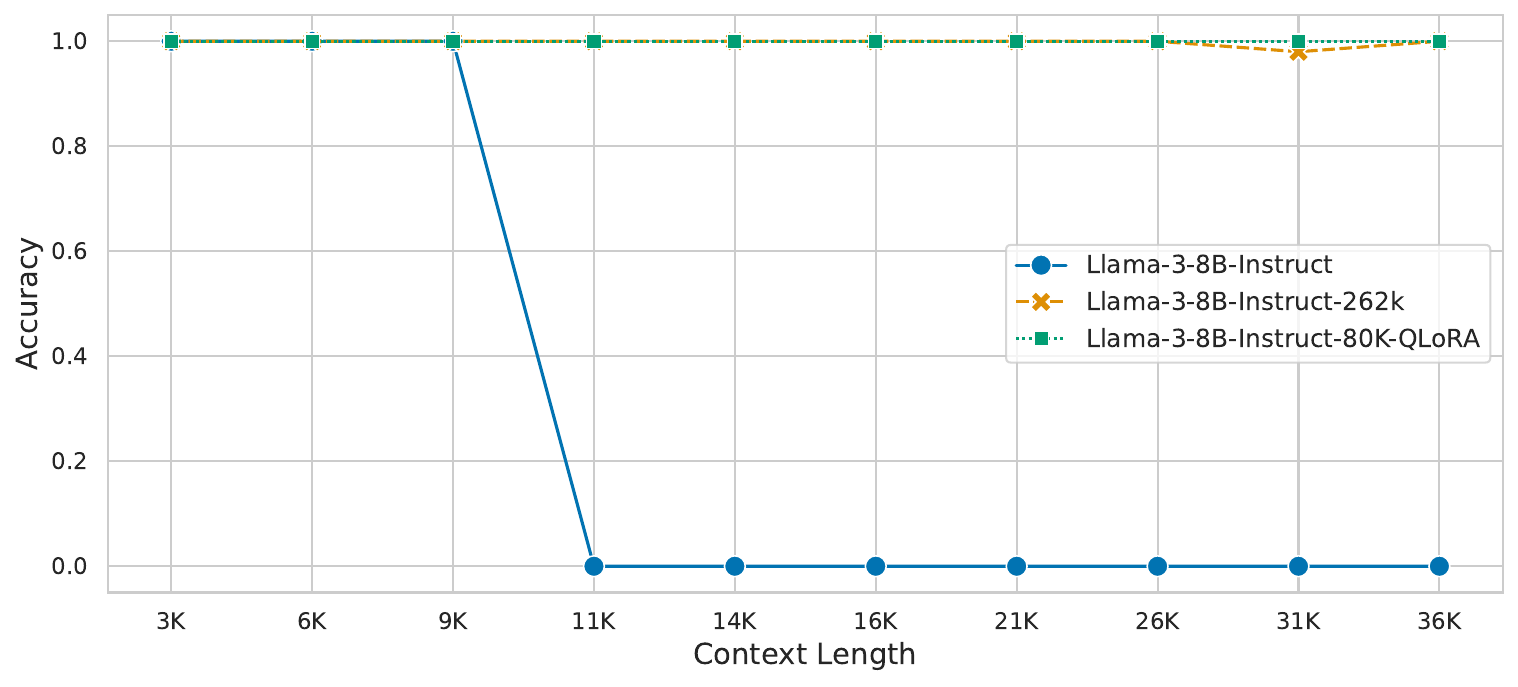}
    \caption{The accuracy of Topic Retrieval task.}
    \label{fig:topic}
\end{figure}

\begin{table}[!t]
    \centering
    \scriptsize
    \begin{tabular}{cccccccc}
        \toprule
        Model & Single-Doc & Multi-Doc & Summ. & Few-Shot & Synthetic & Code & Avg \\
        \midrule
        Llama-3-8B-Instruct & 37.33 & 36.04 & 26.83 & \textbf{69.56} & 37.75 & 53.24 & 43.20 \\
        Llama-3-8B-Instruct-262K & 37.29 & 31.20 & 26.18 & 67.25 & 44.25 & \textbf{62.71} & 43.73 \\
        Llama-3-8B-Instruct-80K-QLoRA & \textbf{43.57} & \textbf{43.07} & \textbf{28.93} & 69.15 & \textbf{48.50} & 51.95 & \textbf{47.19}\\
        \bottomrule
    \end{tabular}
    \caption{Evaluation results on LongBench. For Llama-3-8B-Instruct, we use 8K context length.}
    \label{tab:longbench}
\end{table}

\begin{table}[!t]
    \centering
    \begin{tabular}{cccc}
        \toprule
        Model & LongBookQA Eng & LongBookSum Eng \\
        \midrule
        GPT-4 & 22.22 & 14.73 \\
        Llama-3-8B-Instruct & 7.00 & \textbf{16.40} \\
        Llama-3-8B-Instruct-262K & 20.30 & 10.34 \\
        Llama-3-8B-Instruct-80K-QLoRA & \textbf{30.92} & 14.73 \\
        \bottomrule
    \end{tabular}
    \caption{Evaluation results on InfBench. For Llama-3-8B-Instruct, we use 8K context length. The results of GPT-4 is copied from the paper~\cite{zhang2024inftybench}.}
    \label{tab:infbench}
\end{table}

\begin{table}[!t]
    \centering
    \begin{tabular}{cccccc}
        \toprule
        Model & STEM & Social & Humanities & Others & Avg \\
        \midrule
        Llama-2-7B-Chat & 35.92 & 54.37 & 51.74 & 51.42 & 47.22 \\
        Mistral-7B-v0.2-Instruct & 48.79 & 69.95 & 64.99 & 61.64 & 60.10 \\
        Llama-3-8B-Instruct & \textbf{53.87} & \textbf{75.66} & \textbf{69.44} & 69.75 & \textbf{65.91} \\
        Llama-3-8B-Instruct-262K & 52.10 & 73.26 & 67.15 & \textbf{69.80} & 64.34 \\
        Llama-3-8B-Instruct-80K-QLoRA & 53.10 & 73.24 & 67.32 & 68.79 & 64.44 \\
        \bottomrule
    \end{tabular}
    \caption{Zero-shot performance on MMLU.}
    \label{tab:mmlu}
\end{table}

We evaluate our model on popular long-context benchmarks, then compare it with the original Llama-3-8B-Instruct model and the long-context Llama-3-8B-Instruct-262K from the community\footnote{\url{https://huggingface.co/gradientai/Llama-3-8B-Instruct-262k}}.

Firstly, we leverage the Needle-In-A-Haystack task, which aims to recall an irrelevant piece of information (a.k.a. needle) inserted into a lengthy context (a.k.a. haystack). The accuracy is evaluated with GPT3.5. We use the same needle and haystack as in the official repository\footnote{\url{https://github.com/gkamradt/LLMTest_NeedleInAHaystack}}. Our model achieves 100\% accuracy over all its training context length. Besides, the model generalizes well to the unseen positions (80K$\sim$128K).

Secondly, we report the Topic Retrieval~\cite{longchat2023} accuracy in Figure~\ref{fig:topic}. This task synthesizes a long conversation with multiple independent discussions of a certain topic between the user and the assistant. Then the LLM is required to repeat the first topic as is in the conversation. We use the conversations made up of [5,10,15,20,25,30,40,50,60,70] topics for evaluation. It can be observed that Llama-3-8B-Instruct fails to remember the topic when the context is longer than 9K. However, the accuracy of our model remains 100\% throughout all context lengths.

Thirdly, we evaluate our model on LongBench~\cite{bai2023longbench}, which contains a variety of real-world long-context tasks. Most context on this benchmark is shorter than 32K. Thus, we use 32K context length by default and 8K for Llama-3-8B-Instruct. The results are shown in Table~\ref{tab:longbench}. Our model significantly and consistently outperforms all baselines except on the code completion task. Mixing more code data in training may mitigate this problem.

Forthly, we employ the English Long-Book QA and the Long-Book Summarization task from InfiniteBench~\cite{zhang2024inftybench} to assess the model's performance on really long context. The testing instances are usually longer than 100K. We truncate them to 80K. According to Table~\ref{tab:infbench}, Llama-3-8B-Instruct-80K-QLoRA excels on answering the questions based on the long context. It also achieves competitive performance against GPT-4 in terms of summarization. Interestingly, Llama-3-8B-Instruct with 8K context outperforms GPT-4 with 128K context on summarization. This is likely to be a metric-oriented issue (currently rouge-f1 is used) since the summary may have different paraphrases, which may not necessarily overlap with the ground truth.

Lastly, in Table~\ref{tab:mmlu}, we compare the zero-shot performance of our model and the baselines on MMLU~\cite{hendrycks2021mmlu} benchmark. We also include Llama-2-7B-Chat~\cite{touvron2023llama2} and Mistral-7B-Instruct-v0.2~\cite{jiang2023mistral} for comparison. It can be observed that both long-context models underperform the original Llama-3-8B-Instruct, indicating that context extension may compromise the model's short-context capability. This observation is in line with previous research~\cite{peng2023yarn}. However, our model's performance is still superior to other open-source models at the same scale.

\bibliography{neurips_2024}
\bibliographystyle{abbrv}

\end{document}